\documentclass{article}

\usepackage{arxiv}

\usepackage[utf8]{inputenc} 
\usepackage[T1]{fontenc}    
\usepackage{hyperref}       
\usepackage{url}            
\usepackage{booktabs}       
\usepackage{amsfonts}       
\usepackage{nicefrac}       
\usepackage{microtype}      
\usepackage{lipsum}
\usepackage{graphicx}
\graphicspath{ {./images/} }

\makeatletter
\let\cas@beginabstract\abstract      
\let\cas@endabstract  \endabstract   
\makeatother

\usepackage{arabtex}
\usepackage{utf8}
\setcode{utf8}

\makeatletter
\let\abstract   \cas@beginabstract   
\let\endabstract\cas@endabstract     
\makeatother

\title{SHAMI-MT: A Syrian Arabic Dialect to Modern Standard Arabic Bidirectional Machine Translation System}

\author{
Serry Sibaee* \\
Prince Sultan University \\
Riyadh - Saudi Arabia \\
\texttt{ssibaee@psu.edu.sa} \\
\And
Omer Nacar* \\
Prince Sultan University \\
Riyadh - Saudi Arabia \\
\texttt{onajar@psu.edu.sa} \\
\And
Yasser Al-Habashi \\
Prince Sultan University \\
Riyadh - Saudi Arabia \\
\texttt{yalhabashi@psu.edu.sa} \\
\AND
Adel Ammar \\
Prince Sultan University \\
Riyadh - Saudi Arabia \\
\texttt{aammar@psu.edu.sa} \\
\And
Wadii Boulila \\
Prince Sultan University \\
Riyadh - Saudi Arabia \\
\texttt{wboulila@psu.edu.sa} \\
}

\begin{document}
\maketitle
\begin{abstract}
The rich linguistic landscape of the Arab world is characterized by a significant gap between Modern Standard Arabic (MSA), the language of formal communication, and the diverse regional dialects used in everyday life. This diglossia presents a formidable challenge for natural language processing, particularly machine translation. This paper introduces \textbf{SHAMI-MT}, a bidirectional machine translation system specifically engineered to bridge the communication gap between MSA and the Syrian dialect. We present two specialized models, one for MSA-to-Shami and another for Shami-to-MSA translation, both built upon the state-of-the-art AraT5v2-base-1024 architecture. The models were fine-tuned on the comprehensive Nabra dataset and rigorously evaluated on unseen data from the MADAR corpus. Our MSA-to-Shami model achieved an outstanding average quality score of \textbf{4.01 out of 5.0} when judged by OPENAI model GPT-4.1, demonstrating its ability to produce translations that are not only accurate but also dialectally authentic. This work provides a crucial, high-fidelity tool for a previously underserved language pair, advancing the field of dialectal Arabic translation and offering significant applications in content localization, cultural heritage, and intercultural communication.
\end{abstract}

\section{Introduction}

The rapid advancement of Natural Language Processing (NLP), driven by Large Language Models (LLMs), has transformed how we interact with information and technology~\cite{Enis2024}. However, the benefits of this revolution have not been distributed evenly across all languages. While high-resource languages like English have seen exponential progress, languages with complex internal diversity, such as Arabic, present unique and persistent challenges~\cite{Saeed2025}. The most prominent of these is the phenomenon of diglossia: the co-existence of a high-variety language, Modern Standard Arabic (MSA), and a spectrum of low-variety, colloquial dialects.

MSA is the language of literature, formal education, news media, and official government business across the Arab world. Yet, it is rarely spoken as a native tongue. Instead, daily life—from casual conversation to social media—is conducted in regional dialects~\cite{Sadat2014}. The Syrian dialect, a major variant within the Levantine Arabic family, is a culturally significant and widely spoken dialect with millions of speakers both within Syria and in the global diaspora. The linguistic distance between MSA and Syrian dialect is substantial, encompassing differences in phonology, vocabulary, grammar, and idiomatic expressions.This linguistic gap renders most MSA-trained NLP models ineffective for understanding or generating authentic dialectal content, creating a digital divide that limits the accessibility of technology for millions of Arabic speakers.

Recent progress in Transformer-based encoder-decoder models has propelled the capabilities of NLP systems across diverse tasks and languages. In this work, we leverage AraT5v2~\cite{nagoudi2022}, a state-of-the-art Arabic text-to-text Transformer model, to address the challenges posed by Syrian dialect processing. AraT5v2 builds upon the T5 architecture~\cite{Raffel2020}, with adaptations specifically tailored for Arabic, making it a strong foundation for dialect-aware tasks. We fine-tune AraT5v2 on newly curated parallel datasets that include both Modern Standard Arabic (MSA) and Syrian Arabic, enabling the model to better handle the linguistic variation caused by diglossia. Importantly, our goal is not to present a state-of-the-art system across all benchmarks, but rather to introduce a flexible, extensible framework that can support a wide range of dialectal Arabic NLP applications. This approach lays the groundwork for future research in bridging the gap between formal and colloquial Arabic varieties in language technologies.

To address this critical gap, we have developed SHAMI-MT, a specialized, bidirectional machine translation system for Modern Standard Arabic (MSA) and the Syrian dialect. As part of this effort, we release two openly accessible models:
\begin{itemize}
\item \textbf{MSA-to-Syrian Dialect}: \href{https://huggingface.co/Omartificial-Intelligence-Space/Shami-MT}{\texttt{huggingface.co/Omartificial-Intelligence-Space/Shami-MT}}
\item \textbf{Syrian Dialect-to-MSA}: \href{https://huggingface.co/Omartificial-Intelligence-Space/SHAMI-MT-2MSA}{\texttt{huggingface.co/Omartificial-Intelligence-Space/SHAMI-MT-2MSA}}
\end{itemize}

These models are part of a growing collection, \textbf{SHAMIYAT}, dedicated to advancing Syrian dialect NLP:
\begin{itemize}
\item \textbf{SHAMIYAT Collection}: \href{https://huggingface.co/collections/Omartificial-Intelligence-Space/shamiyat-a-collection-of-syrian-dialect-datasets-and-llms-688cd992267cf3c6b2156184}{\texttt{huggingface.co/collections/Omartificial-Intelligence-Space/shamiyat...}}
\end{itemize}

Our project makes the following key contributions:

\begin{enumerate}
\item \textbf{A Bidirectional Translation System:} We present two distinct models enabling translation from MSA to Syrian dialect and vice versa, supporting a complete communication cycle.
\item \textbf{Leveraging a Specialized Arabic LLM:} We build on AraT5v2, a Transformer model pre-trained specifically on Arabic, offering a deeper understanding of the language than multilingual alternatives.
\item \textbf{Rigorous and Transparent Evaluation:} We assess model performance using the MADAR corpus and an advanced automated judge (GPT-4.1), reporting both strengths and failure modes.
\item \textbf{A Publicly Available Resource:} By releasing these models and datasets, we aim to foster innovation in the under-resourced domain of Syrian dialectal NLP and encourage broader community collaboration.
\end{enumerate}

\section{Related Work}\label{sec:RW}

The challenge of translating between Dialectal Arabic (DA) and MSA is not new, and the approaches to solving it have evolved in lockstep with broader trends in NLP. The first attempts at DA-MSA translation were dominated by rule-based and statistical methods. Rule-based systems relied on manually crafted linguistic rules, morphological analyzers (e.g., the Buckwalter Morphological Analyzer), and extensive lexicons to map dialectal forms to their MSA counterparts~\cite{bouamor2018}. These systems were labor-intensive and brittle, struggling to cope with the vast, unstandardized nature of dialectal orthography and out-of-vocabulary (OOV) words. Phrase-Based Statistical Machine Translation (SMT), as exemplified by systems like Moses, represented an improvement by learning probabilistic translation patterns from parallel corpora \cite{khalifa2016}. However, the performance of SMT was fundamentally capped by the scarcity of large, high-quality, parallel DA-MSA datasets, a bottleneck that persists to this day for many dialect pairs.

The advent of Neural Machine Translation (NMT), particularly with the introduction of the Transformer architecture, marked a paradigm shift. The ability of encoder-decoder models to learn rich, contextual representations of text allowed them to capture the complex, non-linear relationships between dialects and MSA far more effectively than their predecessors \cite{abdelali2021}. NMT models could learn grammatical transformations and lexical substitutions end-to-end, without the need for explicit, hand-crafted rules. This led to significant improvements in translation fluency and accuracy.

More recently, the landscape has been reshaped by massive, pre-trained Large Language Models (LLMs) such as mBERT~\cite{Libovický2019} and mT5~\cite{Xue2020}. These multilingual models are trained on hundreds of languages simultaneously and have demonstrated impressive zero-shot and few-shot learning capabilities. However, for specialized and linguistically nuanced tasks like dialectal translation, their performance can be limited. Because their representational capacity is distributed across many languages, the quality and depth of learning for any single language—particularly low-resource dialects—may be insufficient. Moreover, their training data for dialects is often sparse, noisy, or inconsistently labeled, which can result in overly generic or "flattened" translations that fail to capture the authentic characteristics of the target dialect~\cite{nagoudi2022}. Several prior studies have explored dialectal Arabic processing, focusing on varieties such as Egyptian, Gulf, and Maghrebi Arabic~\cite{Nacar2024}; however, comparatively little attention has been given to the Syrian dialect, which remains underrepresented in both datasets and model development efforts.

This limitation highlights a growing consensus in the NLP community: for high-fidelity performance on specific languages or domains, specialized models are essential. This led to the development of models like AraT5, which was pre-trained exclusively on a vast corpus of Arabic text. By focusing on a single language family, such models can develop a much deeper and more nuanced understanding of its morphology and syntax. Our work on SHAMI-MT is a direct extension of this philosophy, arguing that further specialization through fine-tuning on a high-quality, dialect-specific dataset is the key to unlocking true, high-fidelity dialectal translation.

\section{Methodology}

Our methodology is grounded in the principle of leveraging a powerful, specialized foundation model and adapting it for a specific, high-value task through targeted fine-tuning. The core of our SHAMI-MT system is the $UBC-NLP/AraT5v2-base-1024$ architecture~\cite{nagoudi2022}. This model is a variant of the Text-to-Text Transfer Transformer (T5) framework, which reframes all NLP tasks as a text-to-text problem. In an encoder-decoder model like T5, the encoder processes the source text to create a rich, numerical representation of its meaning. The decoder then uses this representation to auto-regressively generate the target text, one token at a time.

We selected AraT5v2 for its distinct advantages in the context of Arabic NLP. Unlike general-purpose multilingual models, AraT5v2 is pre-trained exclusively on a large and diverse corpus of high-quality Arabic text, granting it a deep understanding of the language’s rich morphology, syntax, and semantics—an essential foundation for any downstream task involving Arabic. Its support for extended sequence lengths, up to 1024 tokens, is particularly valuable for machine translation, as it enables the model to maintain coherence and context across long, syntactically complex sentences that are common in both formal MSA and colloquial dialects. Additionally, AraT5v2 is optimized for fine-tuning, demonstrating efficient convergence and training stability across a variety of tasks. Finally, its encoder-decoder architecture is purpose-built for text generation, making it especially well-suited for translation tasks that require both linguistic accuracy and natural fluency.

\begin{figure}[h!]
\centering
\includegraphics[width=0.8\textwidth]{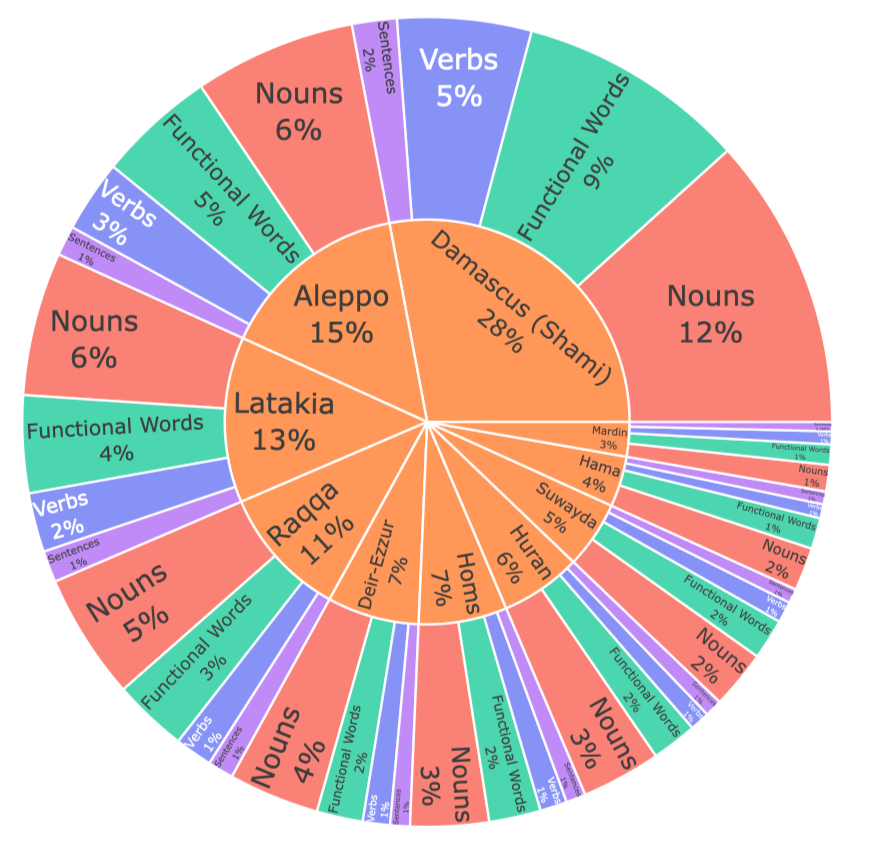}
\caption{The Nâbra Dataset provides a rich and diverse collection of Syrian Arabic from various sources and regions, crucial for training a nuanced dialectal model.}
\label{fig:data}
\end{figure}

To teach our model the specifics of the Syrian dialect, we fine-tuned it on the Nâbra dataset \cite{nayouf2023nabra}. This corpus is an invaluable resource due to its authenticity and diversity. It is not a sterile, academic dataset; rather, it is compiled from a wide array of real-world sources, including social media posts, film and television scripts, song lyrics, and traditional proverbs. Furthermore, it covers a broad geographical range of Syrian sub-dialects, including those from Damascus, Aleppo, Homs, Latakia, and more. This richness ensured that our model was exposed to authentic, varied, and contemporary usage of the Shami dialect. Figure~\ref{fig:data} shows the richness and diversity collection of Syrian Arabic from various sources and regions, crucial for training a nuanced dialectal model.

For a rigorous and impartial assessment of our model's performance, we used a blind test set sourced from the MADAR (Multi-Arabic Dialect Applications and Resources) parallel corpus~\cite{bouamor2018}. We specifically selected 1,500 sentence pairs from the Damascus dialect subset. MADAR is widely recognized as a gold-standard benchmark in the field of Arabic dialectology and NLP. Using it for evaluation ensures that our results are comparable, reproducible, and tested against a standard measure of quality.

The fine-tuning process was carefully designed to balance effective learning with model stability. Training was conducted over 22 epochs, with a total of 10,384 steps, using a batch size of 256. A cosine learning rate schedule was employed, beginning with an initial learning rate of 5e-5. This scheduling strategy allows for a gradual reduction in the learning rate, helping the model converge more smoothly and avoid overshooting minima during optimization. By the end of training, the model achieved a final training loss of 1.396 and a final evaluation loss of 0.771, indicating a well-calibrated fine-tuning process that preserved generalization while improving task-specific performance.

\section{Evaluation and Results}

To move beyond simple lexical metrics like BLEU, which often fail to capture semantic and dialectal nuance, we designed a more sophisticated evaluation framework. We leveraged the advanced reasoning capabilities of GPT-4.1 as an automated evaluator. For each of the 1,500 test sentences from the MADAR corpus, GPT-4.1 was given the MSA input, the model's predicted Shami translation, and the ground truth Shami translation. It was then prompted to provide a holistic quality score from 0 to 5 based on three explicit criteria:
\begin{enumerate}
\item \textbf{Semantic Accuracy:} Does the translation preserve the core meaning of the source sentence?
\item \textbf{Dialectal Authenticity:} Does the translation use vocabulary, grammar, and idioms that are natural for the Syrian dialect?
\item \textbf{Fluency:} Is the translation grammatically correct and easy to read?
\end{enumerate}

\begin{figure}[t]
\centering
\includegraphics[width=0.9\textwidth]{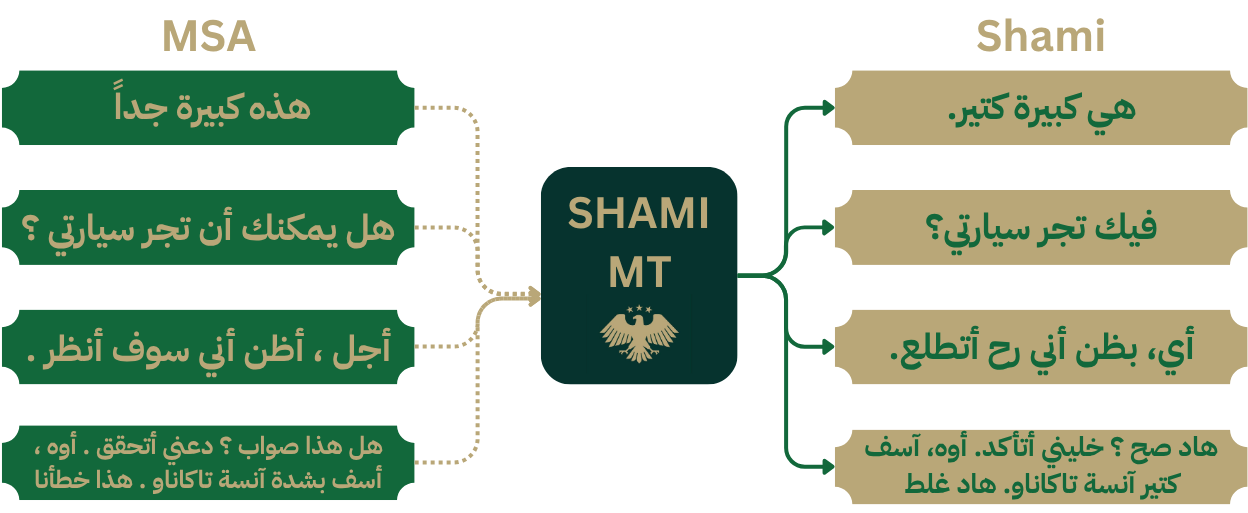}
\caption{Examples of bidirectional translation using SHAMI-MT}
\label{fig:model_examples}
\end{figure}

Across the entire blind test set, the SHAMI-MT model achieved an outstanding average score of 4.01 out of 5.0. This score places the model's average performance squarely in the "very good" to "excellent" range, confirming its ability to consistently produce high-quality, reliable translations. To demonstrate the practical output of our system, Figure~\ref{fig:model_examples} shows some of the SHAMI-MT’s translation performance between MSA and Syrian Arabic.

As shown in~\ref{fig:model_examples}, the system accurately captures both formal and colloquial expressions, preserving meaning while adapting vocabulary, structure, and tone to the target variety.

While aggregate scores provide a general overview of model performance, they often fail to capture the full complexity of translation quality—especially in tasks involving colloquial language. To gain deeper insight into the system’s behavior, we conducted a detailed qualitative analysis of model outputs, evaluating both high-performing and low-performing examples.

The SHAMI-MT system consistently delivered high-quality translations, particularly in cases requiring contextual understanding and idiomatic fluency. In many examples, the model went beyond simple word-level substitution to produce fluid and natural outputs in the Syrian dialect. Table~\ref{tab:evaluation-good} presents representative samples from the MADAR test corpus, illustrating successful MSA-to-Shami translations.

\begin{table}[!ht]
\centering
\caption{Examples of High-Scoring Translations on the MADAR Corpus}
\label{tab:evaluation-good}
\begin{tabular}{|p{3.2cm}|p{3.2cm}|p{3.2cm}|c|p{3.8cm}|}
\hline
\textbf{MSA Input} & \textbf{Model Prediction} & \textbf{Ground Truth (Shami)} & \textbf{Score} & \textbf{Comment} \\
\hline
\RL{استمر في السير في هذا الطريق حتى تجد صيدلية .} & 
\RL{ضل ماشي بهالطريق لتلاقي صيدلية.} & 
\RL{إمشي مباشرة لحد ما تشوف صيدلية.} & 
5 & 
Accurately conveys the intended meaning using natural dialectal expressions. Both model output and reference are idiomatic and correct. \\
\hline
\RL{كيف أستطيع مساعدتك ؟} & 
\RL{كيف فيني ساعدك؟} & 
\RL{كيف فيني ساعدك؟} & 
5 & 
Perfect match with ground truth; fluent and semantically precise. \\
\hline
\RL{اتجه يسارا عند الناصية الثالثة .} & 
\RL{اتجه عاليسار عند الناصية التالتة.} & 
\RL{لف عالشمال بالدخلة التالتة.} & 
4 & 
Very natural phrasing. Minor lexical variation between model and reference, but both are valid Shami translations. \\
\hline
\end{tabular}
\end{table}

On the other hand, a thorough evaluation also requires acknowledging instances where the model underperformed. In Table~\ref{tab:evaluation-bad}, we examine low-scoring examples that highlight two primary sources of error: (1) inconsistencies or mismatches within the evaluation dataset, and (2) legitimate limitations in the model’s ability to handle highly idiomatic or minimal-context expressions.

\begin{table}[ht]
\centering
\caption{Analysis of Representative Low-Scoring Translations}
\label{tab:evaluation-bad}
\begin{tabular}{|p{3.2cm}|p{3.2cm}|p{3.2cm}|c|p{3.8cm}|}
\hline
\textbf{MSA Input} & \textbf{Model Prediction} & \textbf{Ground Truth (Shami)} & \textbf{Score} & \textbf{Comment} \\
\hline
\RL{متى سنستقل القارب ؟} & 
\RL{أيمت رح نستقل القارب؟} & 
\RL{من وين منركب بالقارب؟} & 
2 & 
Model preserves the temporal aspect of the original question, while the ground truth shifts to a spatial focus. Both are plausible interpretations, but differ in intent. \\
\hline
\RL{السيارة جاءت لي من الجانب عند التقاطع .} & 
\RL{السيارة إجت لعندي من الجهة عند التقاطع.} & 
\RL{خبطتني السيارة من عجنب عالمفرق.} & 
2 & 
The model produces a faithful, literal translation. The reference uses more colloquial and compressed phrasing. Semantic alignment exists, but stylistic divergence lowers the score. \\
\hline
\RL{هاهي .} & 
\RL{هي هيي.} & 
\RL{ليكه.} & 
1 & 
The model opts for a more formal repetition, while the reference uses a highly idiomatic single-word Shami expression. Highlights challenges in handling ultra-short, highly informal inputs. \\
\hline
\end{tabular}
\end{table}

This analysis reinforces two key observations. First, the model often produces valid translations that diverge stylistically or semantically from the reference, reflecting the inherent subjectivity and variation in dialectal expression. Second, certain low scores stem from the model's difficulty with brief, highly idiomatic phrases, where surface-level similarity does not guarantee naturalness. These findings suggest that future improvements should target both the breadth of stylistic variation in training data and the refinement of evaluation methods for subjective tasks like dialect translation.

\section{Applications and Future Directions}

The successful development of the SHAMI-MT system opens the door to a wide range of immediate and impactful applications that were previously impractical.
\begin{itemize}
\item \textbf{Content Localization}: The most direct application is in media and technology. Companies can now accurately translate movie subtitles, localize software interfaces and mobile apps, and adapt marketing materials to resonate with a Syrian audience, moving beyond generic MSA that can feel stiff and unnatural.
\item \textbf{Cultural Preservation and Linguistics}: For researchers, the models serve as a powerful tool. Linguists can use the Shami-to-MSA model to standardize dialectal texts for easier analysis, while historians and sociologists can process large volumes of dialectal social media data to understand cultural trends.
\item \textbf{Educational Technology}: Language learning platforms can integrate SHAMI-MT to create dynamic educational tools. An Arabic learner could input an MSA sentence and see its authentic Syrian equivalent, complete with explanations, bridging the critical gap between textbook Arabic and real-world communication.
\item \textbf{Enhanced Intercultural Communication}: The models can facilitate clearer communication in various social and professional settings, from customer service chatbots that can understand dialectal queries to tools that help aid workers and diplomats better understand local contexts.
\end{itemize}

Looking forward, we have identified several promising avenues for future research. The immediate next step is to expand the system's capabilities to other major Levantine dialects, such as Lebanese, Palestinian, and Jordanian, to explore the potential for cross-dialectal transfer learning. Furthermore, we plan to investigate more efficient fine-tuning techniques, such as Low-Rank Adaptation (LoRA), to reduce the computational cost of adapting the model. Finally, integrating these text models with speech recognition and synthesis systems could pave the way for a comprehensive, real-time speech-to-speech translation system for Arabic dialects.

\section{Conclusion}
This paper has detailed the development and evaluation of SHAMI-MT and SHAMI-MT-2MSA, a pair of high-fidelity, bidirectional machine translation models for Modern Standard Arabic and the Syrian dialect. By leveraging a state-of-the-art, Arabic-specific transformer architecture (AraT5v2) with targeted fine-tuning on a rich, authentic dialectal corpus (Nâbra), we have created a system that produces translations of exceptional quality, accuracy, and dialectal authenticity. Our rigorous evaluation on the MADAR benchmark confirms the model's robust performance.

SHAMI-MT successfully fills a critical void in the Arabic NLP landscape, providing an invaluable resource for applications ranging from content localization to cultural preservation. More broadly, our work underscores the importance of specialization in an era of massive language models, demonstrating that for the nuanced and complex challenge of dialectal translation, a focused approach yields superior results. We believe this work sets a new standard for dialectal machine translation and will serve as a catalyst for further innovation in this vital and under-resourced field.

\section*{Acknowledgments}

We would like to thank Prince Sultan University for their generous support in enabling this research.

\bibliographystyle{unsrt}

\end{document}